\documentclass{article}

\usepackage[preprint, nonatbib]{neurips_2025}

\usepackage[utf8]{inputenc} 
\usepackage[T1]{fontenc}    
\usepackage{hyperref}       
\usepackage{url}            
\usepackage{booktabs}       
\usepackage{amsfonts}       
\usepackage{nicefrac}       
\usepackage{microtype}      
\usepackage{multirow}
\usepackage[table]{xcolor}
\usepackage{colortbl}
\usepackage{pifont}
\usepackage{siunitx}
\usepackage{pdfpages}
\usepackage{makecell}
\usepackage{graphicx}
\usepackage{longtable}
\usepackage{float}
\usepackage{tcolorbox}
\usepackage{caption}
\usepackage{placeins}
\title{MedSG-Bench: A Benchmark for Medical Image Sequences Grounding}

\author{%
Jingkun Yue\textsuperscript{1}\quad
Siqi Zhang\textsuperscript{1}\quad
Zinan Jia\textsuperscript{1}\quad
Huihuan Xu\textsuperscript{1}\quad \\
\textbf{Zongbo Han}\textsuperscript{\textbf{2}}\quad
\textbf{Xiaohong Liu}\textsuperscript{\textbf{3}}\quad
\textbf{Guangyu Wang}\textsuperscript{\textbf{1}}\thanks{Corresponding author.}  \quad 
\\
\\
\textsuperscript{1}Beijing University of Posts and Telecommunications\quad 
\textsuperscript{2}Tianjin University\quad \\
\textsuperscript{3}South China Hospital, Medical School, Shenzhen University
}

\begin{document}

\maketitle

\begin{abstract}

Visual grounding is essential for precise perception and reasoning in multimodal large language models (MLLMs), especially in medical imaging domains. While existing medical visual grounding benchmarks primarily focus on single-image scenarios, real-world clinical applications often involve sequential images, where accurate lesion localization across different modalities and temporal tracking of disease progression (e.g., pre- vs. post-treatment comparison) require fine-grained cross-image semantic alignment and context-aware reasoning. To remedy the underrepresentation of image sequences in existing medical visual grounding benchmarks, we propose MedSG-Bench, the first benchmark tailored for \textbf{Med}ical Image \textbf{S}equences \textbf{G}rounding. It comprises eight VQA-style tasks, formulated into two paradigms of the grounding tasks, including 1) Image Difference Grounding, which focuses on detecting change regions across images, and 2) Image Consistency Grounding, which emphasizes detection of consistent or shared semantics across sequential images. MedSG-Bench covers 76 public datasets, 10 medical imaging modalities, and a wide spectrum of anatomical structures and diseases, totaling 9,630 question–answer pairs. We benchmark both general-purpose MLLMs (e.g., Qwen2.5-VL) and medical-domain specialized MLLMs (e.g., HuatuoGPT-vision), observing that even the advanced models exhibit substantial limitations in medical sequential grounding tasks. To advance this field, we construct MedSG-188K, a large-scale instruction-tuning dataset tailored for sequential visual grounding, and further develop MedSeq-Grounder, an MLLM designed to facilitate future research on fine-grained understanding across medical sequential images. The benchmark, dataset, and model are available at \url{https://huggingface.co/MedSG-Bench}
\end{abstract}

\section{Introduction}

Visual grounding is the key step that transforms MLLMs from coarse alignment between language expressions and corresponding visual regions to fine-grained visual understanding and reasoning\cite{xiao2024towards}. For example, models like ChatGPT O3\cite{chatgpto3} often first identify image regions relevant to the questions during reasoning, which helps reduce hallucinations and enhances the trustworthiness of the results.
This capability is particularly crucial in medical imaging, where understanding the semantic content of clinical text (e.g., radiology reports) and accurately localizing the corresponding pathological regions is essential for interpretable and reliable diagnosis\cite{zou2024medrg,chen2023medical,bannur2024maira}.

\begin{figure}[]
\centering
\begin{minipage}{.48\linewidth}
    \includegraphics[width=\linewidth]{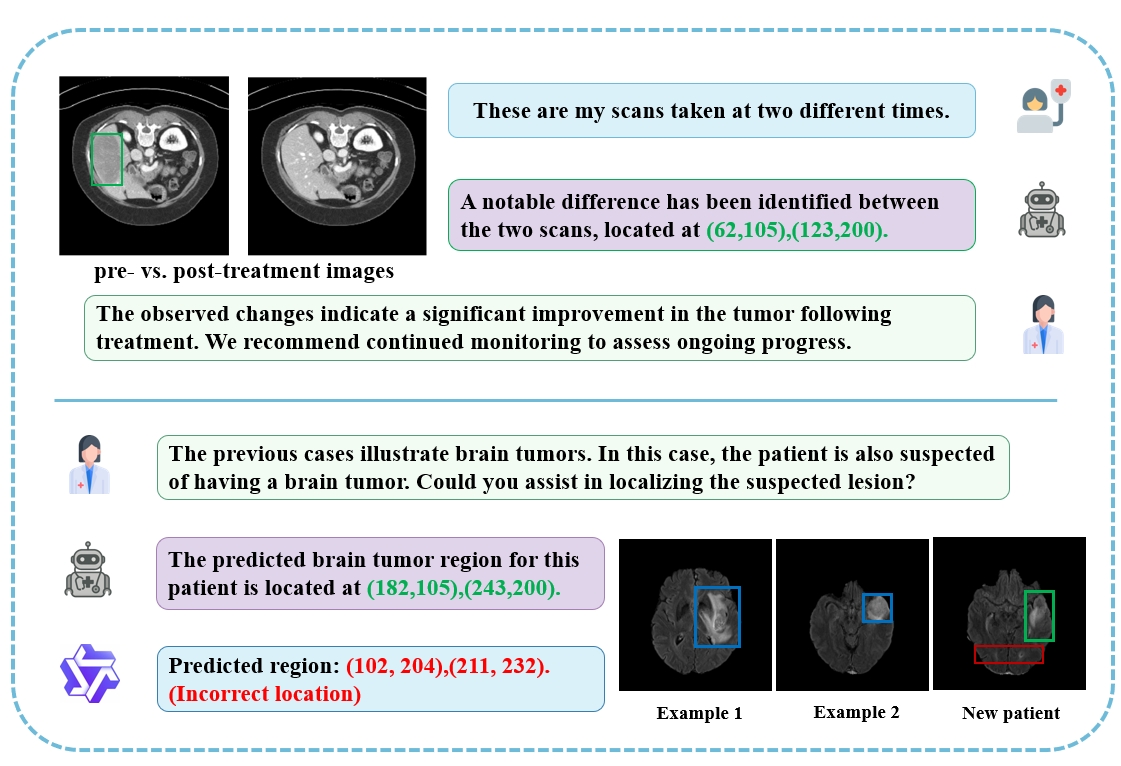}
    \caption{Examples of medical image sequences grounding.}
    \label{fig:fig1}
\end{minipage}
\hfill
\begin{minipage}{.48\linewidth}
    \includegraphics[width=\linewidth]{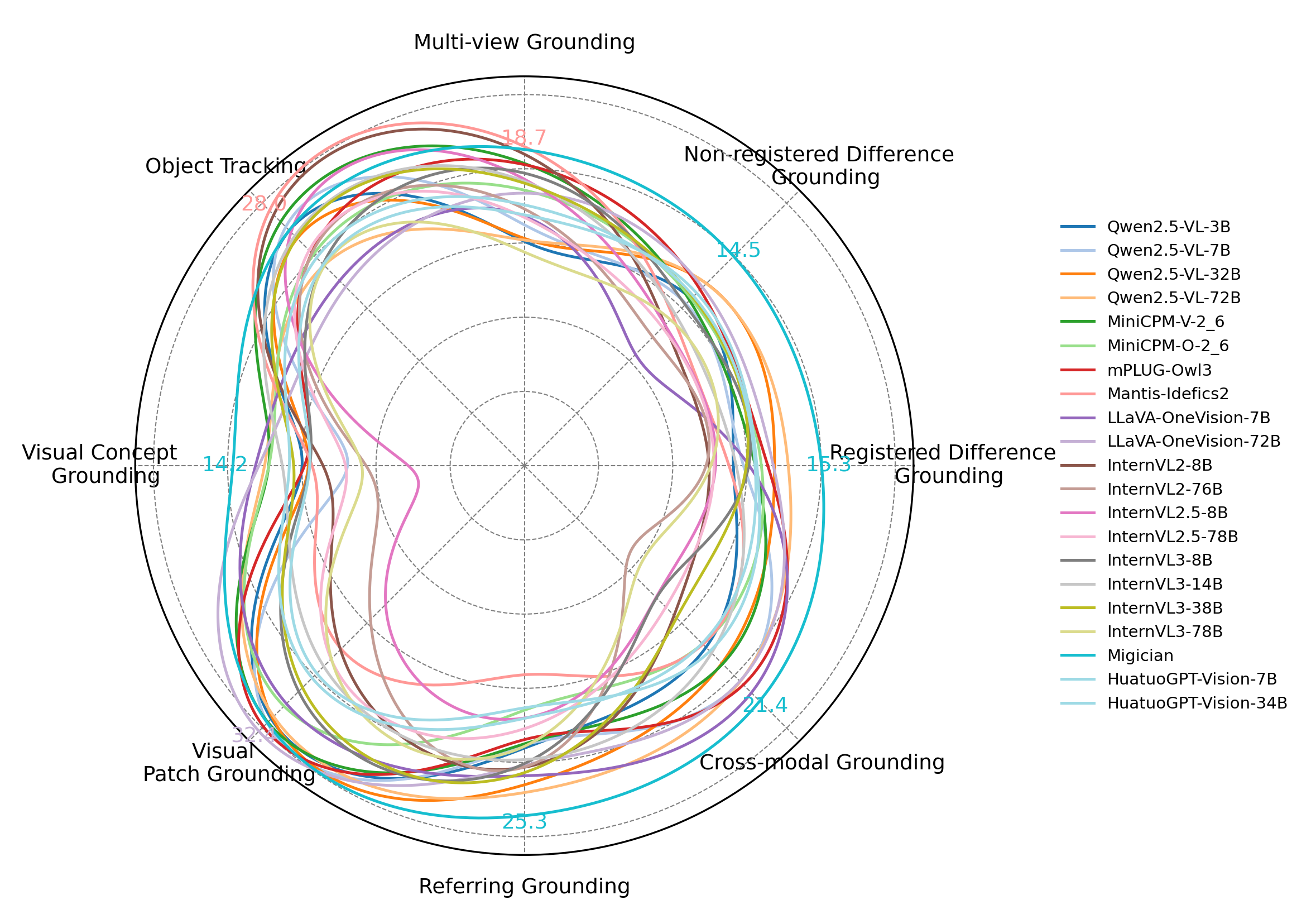}
    \caption{Comparing mainstream MLLMs on MedSG-Bench.}
    \label{fig:fig2}
\end{minipage}

\end{figure}

Currently, existing medical visual grounding benchmarks focus mainly on single-image scenarios \cite{boecking2022making,huang2025towards}. However, real-world clinical diagnosisinherently requires sequential image analysis. As illustrated in Fig.~\ref{fig:fig1}, when assessing disease progression, clinicians routinely perform cross-image comparison (pre- vs. post-treatment images), tracking lesion evolution by analyzing changes in size, morphology, and signal intensity across longitudinal CT scans rather than relying solely on a single static image\cite{mu2025mmxu}. This essential practice of lesion localization and semantic alignment across multiple images forms the cornerstone of reliable clinical reasoning, yet remains underrepresented in current benchmarks.

To address this gap, we introduce MedSG-Bench, the first comprehensive benchmark specifically designed for medical visual grounding in sequential images. Built upon 76 publicly available medical imaging datasets, covering 10 imaging modalities, and 114 clinical tasks, our benchmark systematically evaluates cross-image grounding capability.
Specifically, MedSG-Bench consists of eight carefully designed VQA-style tasks, organized into two grounding paradigms: 1) Image Difference Grounding, which targets the detection of differing regions between sequential images, and 2) Image Consistency Grounding, which focuses on discovering semantically consistent or shared regions across image sequences. This dual-paradigm grounding benchmark can evaluate the essential clinical competencies required for medical image analysis.

In summary, the contributions of this work are as follows:

1. We introduce MedSG-Bench, the first benchmark comprising 9,630 VQA-style samples specifically designed to evaluate the grounding capabilities of MLLMs in medical image sequences. The benchmark defines eight tasks grouped into two core paradigms, Image Difference Grounding and Image Consistency Grounding, which jointly serve to evaluate essential clinical competencies required for medical image analysis.

2. We conduct comprehensive evaluations of both general-purpose MLLMs (e.g., Qwen2.5-VL\cite{bai2025qwen2}) and medical-domain specialized MLLMs (e.g., HuatuoGPT-Vision\cite{chen2024huatuogpt}) on MedSG-Bench. Our results (Fig.~\ref{fig:fig2}) show that all current MLLMs exhibit substantial limitations in fine-grained grounding of medical image sequences. 

3. To promote progress in this underexplored area, we construct MedSG-188K, a large-scale instruction-tuning dataset tailored for grounding in medical image sequences. Based on this dataset, we further develop MedSeq-Grounder, and achieves state-of-the-art performance on MedSG-Bench.

\section{Related work}
\label{gen_inst}
\subsection{Multimodal Large Language Models}
Recent advances in multimodal large language models (MLLMs) have progressively extended their capabilities from coarse image-level understanding to fine-grained visual grounding\cite{xiao2024towards,krishna2017visual}. This progress has been primarily achieved through three main approaches: 1) instruction tuning with grounding supervision\cite{chen2023shikra,peng2023kosmos}, 2) integrating external localization modules\cite{lai2024lisa,xia2024gsva,rasheed2024glamm,zhang2024psalm,ren2024pixellm,zhang2023next} such as SAM\cite{kirillov2023segment} or Grounding DINO\cite{liu2024grounding}, and 3) leveraging vision tokenizers to enable perceive-then-understand paradigms\cite{ma2024groma,jiang2024chatrex}. While these methods have significantly improved grounding accuracy within individual images, they largely overlook the clinically relevant and more complex setting of multi-image visual grounding. Migician\cite{li2025migician} is the first model to tackle this challenge in the natural image domain, enabling free-form and accurate grounding across multiple images. Building upon this paradigm, we extend the exploration to the medical domain, focusing on sequential visual grounding in clinically meaningful scenarios.

\subsection{Medical MLLM Benchmarks}
As shown in Table~\ref{tab:comparison_exist_benchmark}, benchmarks in the medical domain have progressed from early settings involving single-image and single-modality inputs to more advanced configurations covering multiple organs\cite{liu2021slake}, cross-modal scenarios\cite{ye2024gmai}, and multi-image understanding\cite{hu2023medical,mu2025mmxu}. Some recent benchmarks have also provided fine-grained annotations to enrich evaluation. However, these benchmarks primarily emphasize image-level understanding. Even when detailed annotations are available, they are typically utilized for classification or question answering tasks, rather than for explicit visual grounding. In contrast, grounding-oriented benchmarks remain scarce in the medical domain and are currently limited to single-image scenarios\cite{boecking2022making,huang2025towards}. To date, no medical benchmark has systematically explored sequential visual grounding, a capability that is essential for various clinical tasks such as cross-view lesion comparison, longitudinal disease progression tracking, and multi-phase imaging interpretation. To fill this gap, we propose MedSG-Bench, the first benchmark dedicated to fine-grained visual grounding in sequential medical images.

\newcommand{\cmark}{\textcolor{green!60!black}{\ding{51}}}  
\newcommand{\xmark}{\textcolor{red}{\ding{55}}}             

\begin{table*}[t]
\caption{Comparison between MedSG-Bench and other existing benchmarks in the medical field. FG denotes fine-grained annotation. $^{*}$ indicates the test set.}

\resizebox{1.0\textwidth}{!}{
\begin{tabular}{l|ccccccc}
\hline
\textbf{Benchmark} & \textbf{Size} & \textbf{Task} & \textbf{Multi-modality} & \textbf{Multi-organ} & \textbf{Image-Sequence} & \textbf{FG} & \textbf{Max Length} \\ \hline

        \rowcolor[RGB]{234, 238, 234}
        \multicolumn{8}{c}{\bf Understanding-oriented medical benchmarks} \\
        \midrule
VQA-RAD\cite{lau2018dataset} & 3K & 11 & \cmark & \cmark & \xmark & \xmark & 1 \\
SLAKE$^{*}$\cite{liu2021slake} & 2K & 10 & \cmark & \cmark & \xmark & \cmark & 1 \\
OmniMedVQA\cite{hu2024omnimedvqa} & 128K & 5 & \cmark & \cmark & \xmark & \xmark & 1\\
GMAI-MMBench\cite{ye2024gmai} & 26K & 18 & \cmark & \cmark & \xmark & \cmark & 1\\
Medical-Diff-VQA$^{*}$\cite{hu2023medical} & 70K & 7 & \xmark & \xmark & \cmark & \xmark & 2 \\
MMXU$^{*}$\cite{mu2025mmxu} & 3K & 3 & \xmark & \xmark & \cmark & \cmark & 2 \\
        \midrule
        \rowcolor[RGB]{234, 238, 234}
        \multicolumn{8}{c}{\bf Grounding-oriented medical benchmarks} \\
        \midrule
MS-CXR$^{*}$\cite{boecking2022making} & 1K & 1 & \xmark & \xmark & \xmark & \cmark & 1 \\
MeCoVQA-G$^{*}$\cite{huang2025towards} & 2K & 1 & \cmark & \cmark & \xmark & \cmark & 1 \\
\hline
MedSG-Bench & 9K & 8 & \cmark & \cmark & \cmark & \cmark & 6 \\ \hline
\end{tabular}
\label{tab:comparison_exist_benchmark}
}
\vspace{-0.4cm}
\end{table*}

\section{MedSG-Bench}

In this section, we provide an in-depth overview of the careful design and development of MedSG-Bench, covering the rigorous collection and preprocessing of medical data, the systematic definition of tasks tailored for sequential visual grounding, and the presentation of detailed dataset statistics.

\subsection{Data Collection and Preprocessing}

\subsubsection{Dataset Review and Selection} As shown in Fig.~\ref{fig:datapro}, open data repositories, including Zenodo, Github, among others, were searched for medical image datasets. Data with permissive licenses (e.g., CC BY 4.0) that allow derivative works and redistribution were given priority during selection. We retained only those datasets that provided local annotations, such as segmentation masks or bounding boxes, which are essential for grounding-based tasks. To ensure mutual exclusivity among imaging cases, we cross-referenced dataset metadata and associated papers to identify and remove duplicated samples. Additionally, we performed a manual quality review to exclude images with poor visual clarity or unreliable annotations, thereby preserving the overall integrity and usability of the data.

\subsubsection{Standardization} Medical imaging datasets exhibit high heterogeneity in format, resolution, intensity distribution, and metadata quality, with modality-specific characteristics that differ markedly from natural images. To mitigate this variability, we followed the preprocessing strategy proposed in \cite{ma2024segment}, applying min-max normalization to rescale pixel intensities to a standardized range, thereby enabling more consistent downstream processing. To unify the data format, both 3D volumetric scans and video sequences were converted into 2D RGB images—achieved by slicing along anatomical axes or sampling frames at fixed intervals, respectively. All images were subsequently resized to 336×336 pixels, and each image was assigned a unique identifier encoding its imaging modality and associated task. Finally, all processed images were stored in lossless PNG format to preserve visual fidelity.

\begin{figure}[t!]
  \centering
  \includegraphics[width=\textwidth]{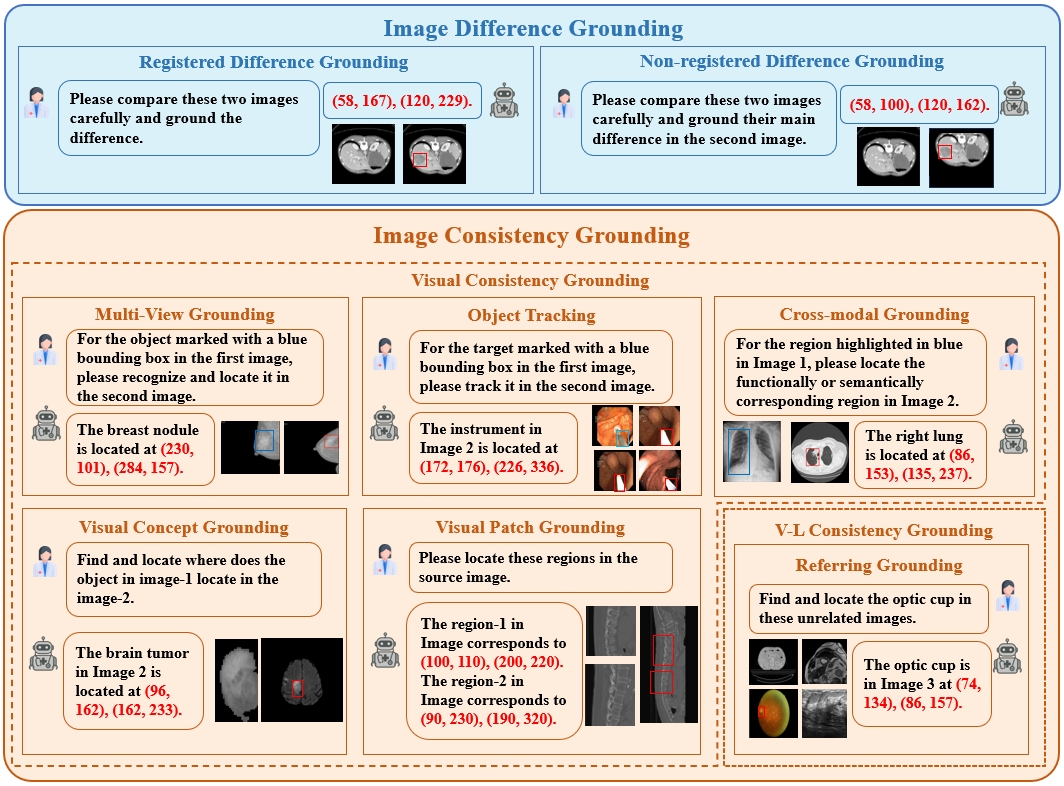}
  \caption{An illustration of medical image sequences grounding tasks included in MedSG-Bench.}
  \label{fig:main}
\end{figure}
\label{headings}

\subsection{VQA tasks definition and generation}

To facilitate fine-grained evaluation of visual grounding for sequential medical images, we define eight VQA-style tasks, organized into two complementary categories, including Image Difference Grounding and Image Consistency Grounding, which collectively capture both semantic changes and invariant features across image sequences, as illustrated in Fig.~\ref{fig:main}.

\subsubsection{Image Difference Grounding}

Image Difference Grounding focuses on detecting and localizing regions of changes across sequential images, enabling assessment of a model's ability to perceive subtle or clinically relevant variations.

\paragraph{Task 1: Registered Difference Grounding}

Given a pair of spatially aligned (i.e., registered) images that are visually identical except for a single region, the model is designed to detect and localize the difference. To generate such image pairs in a controlled and scalable manner, we begin with a single medical image and introduce localized perturbations that simulate clinically meaningful variations, such as disease progression or treatment response. These perturbations comprise both geometric or appearance-based transformations (e.g., CutPaste\cite{li2021cutpaste}), and synthetic anomalies generated using state-of-the-art medical generative models\cite{wu2025freetumor,hu2023label,chen2024towards}. To avoid the model learning shortcuts, such as associating a fixed image position with abnormalities, we randomize the ordering of image pairs, ensuring that either the normal or the abnormal image may appear in either position.

\paragraph{Task 2: Non-registered Difference Grounding}

In clinical practice, medical images often exhibit spatial misalignments due to patient movement, scanner variability, or imperfect registration. This issue is particularly common when comparing medical images acquired from the same patient at different time points, where the lack of proper registration can lead to spatial shifts in organs or lesions, thereby potentially challenging models to distinguish real differences from registration artifacts.
To better simulate such conditions and evaluate the model's robustness to Non-registered Difference Grounding, we extend Task 1 by introducing controlled spatial shifts: each image is randomly translated by up to 20 pixels along both the horizontal and vertical axes. The model is thus required to identify and accurately localize the primary difference between the two images while ignoring changes caused by misalignment.

\begin{figure}[t!]
  \centering
  \includegraphics[width=\textwidth]{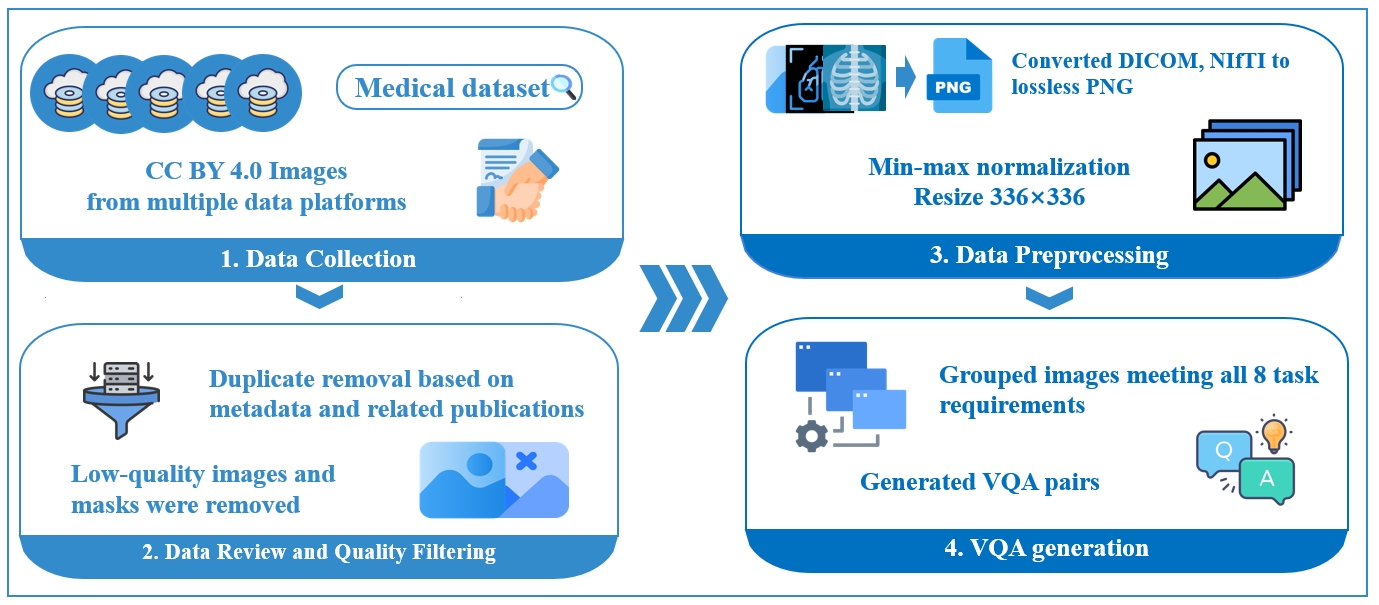}
  \caption{Overview of the MedSG-Bench construction protocol.}
  \label{fig:datapro}
\end{figure}
\subsubsection{Image Consistency Grounding}

Image Consistency Grounding focuses on identifying and aligning invariant semantics across sequential medical images, which is essential for cross-view, cross-modal and cross-time alignment in clinical practice.

Specifically, Image Consistency Grounding can be divided into two subcategories: 1) Visual Consistency Grounding (Task 3-7), which evaluates the model’s ability to capture visual consistency across multiple images; 2) Vision-Language Consistency Grounding (Task 8), which involves aligning language-referenced information with multiple medical images.

\paragraph{Task 3: Multi-View Grounding}
Medical images from different views often have geometric inconsistencies due to patient movement, scanning protocols, or anatomical deformation.

To assess a model's ability to capture cross-view correspondence, we construct the Multi-View Grounding task using two implementation strategies. First, we repurpose existing multi-view datasets (e.g., VinDr-Mammo) by converting them into a VQA-style format. Second, we simulate multi-view scenarios by extracting three orthogonal slices (axial, sagittal, and coronal) from 3D medical volumes. Notably, the reference view is not fixed and may vary across different samples.

\paragraph{Task 4: Object Tracking}
Accurately tracking anatomical structures or instruments across slices of medical images or frames of surgical video  is essential in clinical workflows (e.g., lesion monitoring and intraoperative navigation).
This task evaluates the model’s ability to maintain consistent localization of a target object across sequential frames or slices. We construct this task using two types of data sources. First, we leverage existing surgical videos, where objects such as instruments or tissues are manually annotated across frames. Second, we simulate spatial tracking scenarios by slicing 3D medical volumes along a fixed anatomical axis, treating anatomical structures or lesions as trackable targets across ordered 2D slices.

\paragraph{Task 5: Visual Concept Grounding}
In clinical scenarios, lesions can exhibit high variability in locations (e.g., across anatomical regions) and visual appearance due to imaging protocols or disease subtypes.
This variability challenges models to learn robust target representations based on pathological features, rather than over-relying on spatial biases.
This task evaluates the model’s ability to recognize and localize a visually distinct and semantically coherent concept, including both pathological findings such as tumors and anatomical structures such as organs or tissue subtypes, within a complex medical image. The model is provided with a reference image in which the concept appears under idealized conditions, and must identify the corresponding instance in a target image with greater visual clutter and contextual complexity. To construct this task, the reference concept is extracted from the target using segmentation masks to ensure semantic consistency.
\begin{figure}[!t]
\centering
\begin{minipage}{\linewidth}
\centering
\small
\captionof{table}{Detailed statistics of MedSG-Bench.}
\label{tab:task-stats}
\vspace{1mm}
\begin{tabular}{l|cccc}
    \toprule
    Task & \#Datasets & \#Modalities & \#Clinical Tasks & Max Length \\
    \midrule
    Registered Difference Grounding        & 50 & 10 & 59 & 2\\
    Non-registered Difference Grounding   & 50 & 10 & 58 & 2\\
    Multi-view Grounding  & 30 & 4 & 75 & 3\\
    Object Tracking  & 30 & 4 & 87 & 6\\
    Visual Concept Grounding  & 49 & 10 & 87 & 2\\
    Visual Patch Grounding  & 53 & 10 & 78 & 5\\
    Cross-modal Grounding  & 24 & 4 & 28 & 4\\
    Referring Grounding  & 9 & 8 & 28 & 3\\
    \midrule
    MedSG-Bench  & 76 & 10 & 114 & 6\\
    \bottomrule
\end{tabular}
\end{minipage}
\end{figure}

\begin{figure}[!t]
  \centering
  \includegraphics[width=\textwidth]{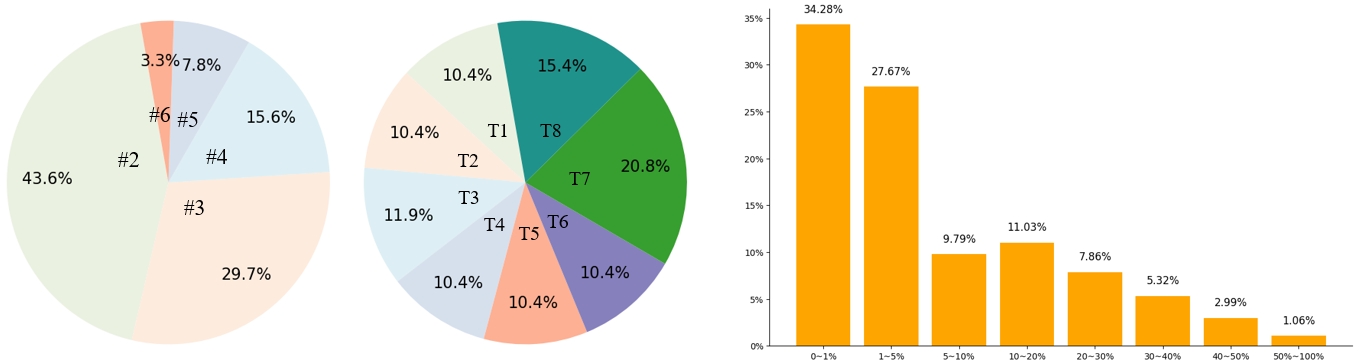}
  \caption{Proportions of image sequence length (\textbf{left}), data distribution across tasks (\textbf{middle}), and target-to-image size ratios (\textbf{right}) in MedSG-Bench.}
  \label{fig:datades1}
\end{figure}

\paragraph{Task 6: Visual Patch Grounding}
Precisely distinguishing nearly identical anatomical structures (e.g., separating tumor margins from adjacent vasculature) is essential for image-guided interventions and radiotherapy planning, where  subtle visual distinctions determine procedural success. Therefore, we design this task evaluates the model’s ability to match a local image patch to its original location within a larger image. It poses significant challenges in contexts where structures like vertebral segments (e.g., T1 to T12) exhibit nearly identical appearances. To construct this task, we initially sample 15 patches per image and manually select up to five based on foreground richness, including organ boundaries, lesion areas, or diagnostically relevant fine structures. The rest are discarded. This selective sampling ensures that each retained patch presents a non-trivial grounding challenge while avoiding visually homogeneous regions.

\paragraph{Task 7: Cross-modal Grounding}
In clinical practice, the same patient is often examined using different imaging modalities such as CT, X-ray, or MRI, each highlighting distinct but complementary aspects of anatomical structures or pathologies.
This task assesses the model’s ability to ground semantically or functionally equivalent regions across differing imaging contexts. Given a reference region from one image, the model is required to identify the corresponding region in a target image that may differ in imaging modality (e.g., CT versus MRI) or contrast type (e.g., T1-weighted versus T2-weighted MRI). Region pairs are manually curated based on metadata such as modality type and annotated labels to ensure semantic alignment and multimodal consistency.

\paragraph{Task 8: Referring Grounding}
Clinicians often describe findings or refer to specific regions using natural language expressions. Enabling models to accurately interpret and associate such expressions with visual content is essential for enhancing interpretability, supporting human-AI collaboration, and building reliable decision support systems.
Considering the prevalence of partially labeled data in medical imaging, we carefully curate candidate image sets to ensure that the images are semantically unrelated. This reduces the risk of referential ambiguity caused by overlapping content or latent correlations among images.

\subsection{Data description}
\label{datades1}
We curated a total of 76 publicly available datasets under permissive licenses, prioritizing those released with open CC-BY terms to ensure broad accessibility. As summarized in Table~\ref{tab:task-stats}, MedSG-Bench spans 10 medical imaging modalities and and encompasses 114 distinct clinical tasks, covering a wide range of anatomical regions and disease types. The benchmark contains 9,630 visual question answering pairs, designed to assess fine-grained grounding capabilities across diverse clinical contexts.
In addition to task coverage, we also provide detailed statistics on the proportion of image sequence lengths, data distribution, and target-to-image size ratios (lesions or anatomical abnormalities are often subtle, localized, and small in size), offering a comprehensive overview of the benchmark’s complexity and representativeness in Fig.~\ref{fig:datades1}.

\section{MedSG-188K and MedSeq-Grounder}

\subsection{MedSG-188K}
The construction of MedSG-188K is based on the eight tasks defined by MedSG-Bench. To ensure diversity in VQA-style queries, we first crafted seed instruction templates tailored to the specific characteristics of each task, capturing the nuanced demands of distinct clinical scenarios. These seed templates were then expanded using GPT-4\cite{achiam2023gpt}, which generated ten diverse free-form instruction variants per task by systematically varying the phrasing, contextual framing, and query structure.
For each medical image sequence, one of the instruction templates was randomly selected and populated with task-specific content to generate diverse question-answer pairs. Using this pipeline, we constructed a total of 188,163 VQA-style samples. The distribution of sequence lengths, data volume is summarized in Fig.~\ref{fig:datades}.

\subsection{MedSeq-Grounder}
\label{grounder}
MedSeq-Grounder is developed based on the Qwen2.5-VL-7B model\cite{bai2025qwen2} and trained using the LLaMA-Factory framework\cite{zheng2024llamafactory}. The training is performed with a global batch size of 64 over 15,000 steps, using a learning rate of 5e-6 and 4×A40-48G GPUs.

\begin{table}[h!] \scriptsize
    \centering
    \caption{Performance of different MLLMs on MedSG-Bench. IDG: Image Difference Grounding; ICG: Image Consistency Grounding; RDG: Registered Difference Grounding; NRDG: Non-registered Difference Grounding; MV: Multi-view Grounding; OT: Object Tracking; VCG: Visual Concept Grounding; VPG: Visual Patch Grounding; CMG: Cross-modal Grounding; RG: Referring Grounding; Avg.: Average; IoU and \textcolor{gray}{acc@0.5} for all results are shown, all numbers are in percentages.}
    \begin{tabular}{lc|cc|cccccc|c}
        \toprule
        \multirow{3}{*}{\bf Model} & \multirow{3}{*}{\bf Size} & \multicolumn{2}{c|}{\bf IDG} & \multicolumn{6}{c|}{\bf ICG} & \multirow{3}{*}{\bf Avg.} \\
        \cmidrule(lr){3-10}
         &  & RDG & NRDG & MV & OT & VCG & VPG & CMG & RG &\\
        \midrule
        \rowcolor[RGB]{234, 238, 234}
        \multicolumn{11}{c}{\bf General-purpose MLLMs} \\
        \midrule
\includegraphics[width=0.4cm]{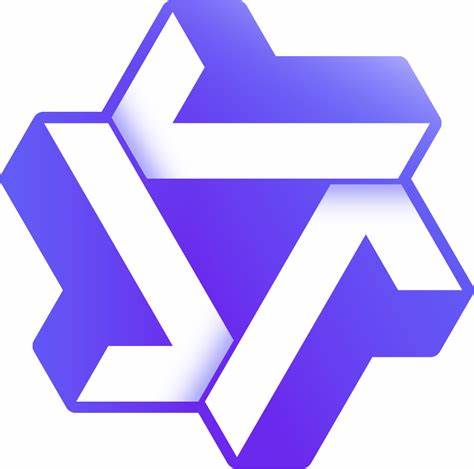} Qwen2.5-VL\cite{bai2025qwen2} & 3B & \makecell{0.59\\\textcolor{gray}{0.30}} & \makecell{1.62\\\textcolor{gray}{1.30}} & \makecell{7.12\\\textcolor{gray}{3.90}} & \makecell{21.32\\\textcolor{gray}{16.80}} & \makecell{6.98\\\textcolor{gray}{0.80}} & \makecell{27.36\\\textcolor{gray}{3.40}} &  \makecell{10.02\\\textcolor{gray}{1.65}} & 
\makecell{12.99\\\textcolor{gray}{6.82}} &
\makecell{10.94\\\textcolor{gray}{4.20}} \\
\cmidrule(lr){2-11}
\includegraphics[width=0.4cm]{Qwen.png} Qwen2.5-VL\cite{bai2025qwen2} & 7B & \makecell{0.88\\\textcolor{gray}{0.30}} & \makecell{1.25\\\textcolor{gray}{0.00}} & \makecell{8.48\\\textcolor{gray}{3.73}} & \makecell{22.41\\\textcolor{gray}{17.80}} & \makecell{4.22\\\textcolor{gray}{1.00}} & \makecell{28.87\\\textcolor{gray}{5.70}} &  \makecell{16.29\\\textcolor{gray}{4.45}} &
\makecell{12.58\\\textcolor{gray}{6.21}} &
\makecell{12.31\\\textcolor{gray}{4.90}} \\
\cmidrule(lr){2-11}
\includegraphics[width=0.4cm]{Qwen.png} Qwen2.5-VL\cite{bai2025qwen2} & 32B & \makecell{2.69\\\textcolor{gray}{1.40}} & \makecell{3.48\\\textcolor{gray}{1.20}} & \makecell{7.35\\\textcolor{gray}{2.61}} & \makecell{19.12\\\textcolor{gray}{13.40}} &  \makecell{6.53\\\textcolor{gray}{1.30}} & \makecell{26.92\\\textcolor{gray}{7.10}} & \makecell{12.59\\\textcolor{gray}{4.90}} &
\makecell{18.71\\\textcolor{gray}{11.67}} & 
\makecell{12.47\\\textcolor{gray}{5.71}} \\
\cmidrule(lr){2-11}
\includegraphics[width=0.4cm]{Qwen.png} Qwen2.5-VL\cite{bai2025qwen2} & 72B & \makecell{4.37\\\textcolor{gray}{2.60}} & \makecell{3.46\\\textcolor{gray}{0.80}} & \makecell{7.22\\\textcolor{gray}{2.78}} & \makecell{13.11\\\textcolor{gray}{7.70}} & \makecell{10.33\\\textcolor{gray}{3.50}} & \makecell{26.45\\\textcolor{gray}{6.30}} &  \makecell{16.32\\\textcolor{gray}{7.00}} & 
\makecell{20.19\\\textcolor{gray}{14.10}} &
\makecell{13.35\\\textcolor{gray}{6.12}} \\
\midrule
\includegraphics[width=0.4cm]{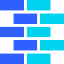} MiniCPM-V-2\_6\cite{yao2024minicpm} & 8B & \makecell{1.36\\\textcolor{gray}{0.00}} & \makecell{1.50\\\textcolor{gray}{0.00}} & \makecell{15.82\\\textcolor{gray}{5.20}} & \makecell{24.03\\\textcolor{gray}{18.50}} & \makecell{9.90\\\textcolor{gray}{2.10}} & \makecell{28.65\\\textcolor{gray}{12.20}} &  \makecell{12.72\\\textcolor{gray}{3.30}} & 
\makecell{12.44\\\textcolor{gray}{3.64}} &
\makecell{13.24\\\textcolor{gray}{5.27}} \\
\cmidrule(lr){2-11}
\includegraphics[width=0.4cm]{minicpm.png} MiniCPM-O-2\_6\cite{minicpm-o} & 8B & \makecell{1.69\\\textcolor{gray}{0.10}} & \makecell{1.63\\\textcolor{gray}{0.00}} & \makecell{12.11\\\textcolor{gray}{2.43}} & \makecell{15.25\\\textcolor{gray}{9.60}} & \makecell{9.88\\\textcolor{gray}{1.70}} & \makecell{22.96\\\textcolor{gray}{9.20}}&  \makecell{9.53\\\textcolor{gray}{2.35}} & 
\makecell{8.82\\\textcolor{gray}{2.02}} &
\makecell{10.12\\\textcolor{gray}{3.23}} \\
\midrule
\includegraphics[width=0.4cm]{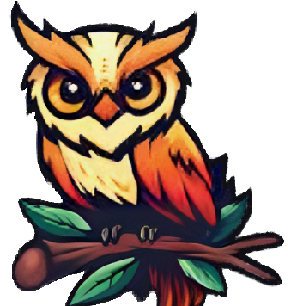} mPLUG-Owl3\cite{ye2024mplug} & 7B & \makecell{2.12\\\textcolor{gray}{0.00}} & \makecell{2.55\\\textcolor{gray}{0.00}} & \makecell{15.64\\\textcolor{gray}{3.64}} & \makecell{15.62\\\textcolor{gray}{4.40}} & \makecell{6.80\\\textcolor{gray}{0.80}} & \makecell{30.42\\\textcolor{gray}{3.60}} &  \makecell{17.06\\\textcolor{gray}{4.80}} & 
\makecell{11.92\\\textcolor{gray}{5.47}} &
\makecell{13.22\\\textcolor{gray}{3.19}} \\
\midrule
\includegraphics[width=0.4cm]{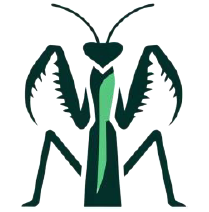} Mantis-Idefics2\cite{jiang2024mantis} & 8B & \makecell{0.49\\\textcolor{gray}{0.00}} & \makecell{0.62\\\textcolor{gray}{0.00}} & \makecell{\underline{18.69}\\\textcolor{gray}{\underline{8.59}}} & \makecell{\underline{28.04}\\\textcolor{gray}{\underline{23.50}}} & \makecell{6.27\\\textcolor{gray}{0.50}} & \makecell{10.26\\\textcolor{gray}{1.10}} & \makecell{9.59\\\textcolor{gray}{0.95}} & 
\makecell{6.05\\\textcolor{gray}{0.54}} & \makecell{9.90\\\textcolor{gray}{3.91}} \\
\midrule        
\includegraphics[width=0.4cm]{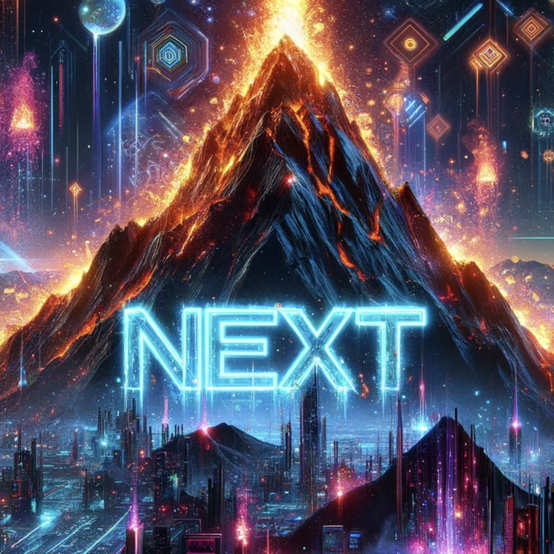} LLaVA-OneVision\cite{li2024llava} & 7B & \makecell{1.09\\\textcolor{gray}{0.00}} & \makecell{0.01\\\textcolor{gray}{0.00}} & \makecell{9.26\\\textcolor{gray}{1.13}} & \makecell{10.50\\\textcolor{gray}{3.20}} & \makecell{11.33\\\textcolor{gray}{1.80}} & \makecell{22.20\\\textcolor{gray}{5.30}} & \makecell{19.08\\\textcolor{gray}{6.70}} & 
\makecell{17.11\\\textcolor{gray}{5.67}} & \makecell{12.39\\\textcolor{gray}{3.47}} \\
\cmidrule(lr){2-11}
\includegraphics[width=0.4cm]{llava.png} LLaVA-OneVision\cite{li2024llava} & 72B & \makecell{2.58\\\textcolor{gray}{0.80}} & \makecell{2.87\\\textcolor{gray}{0.90}} & \makecell{11.74\\\textcolor{gray}{1.39}} & \makecell{9.61\\\textcolor{gray}{2.30}} & \makecell{10.95\\\textcolor{gray}{3.30}} & \makecell{\underline{32.38}\\\textcolor{gray}{\underline{20.30}}} & 
\makecell{16.24\\\textcolor{gray}{5.40}} & \makecell{15.43\\\textcolor{gray}{6.68}} & \makecell{13.21\\\textcolor{gray}{5.18}} \\
\midrule  
\includegraphics[width=0.4cm]{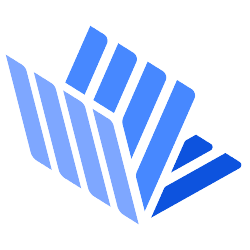} InternVL2\cite{chen2024internvl} & 8B & \makecell{0.18\\\textcolor{gray}{0.00}} & \makecell{0.38\\\textcolor{gray}{0.00}} & \makecell{17.34\\\textcolor{gray}{7.03}} & \makecell{26.45\\\textcolor{gray}{21.20}} & \makecell{5.56\\\textcolor{gray}{0.80}} & \makecell{10.36\\\textcolor{gray}{0.70}} &  \makecell{6.23\\\textcolor{gray}{1.00}} & 
\makecell{15.73\\\textcolor{gray}{7.69}} &
\makecell{10.24\\\textcolor{gray}{4.59}} \\
\cmidrule(lr){2-11}
\includegraphics[width=0.4cm]{internvl.png} InternVL2\cite{chen2024internvl} & 76B & \makecell{0.15\\\textcolor{gray}{0.00}} & \makecell{0.15\\\textcolor{gray}{0.00}} & \makecell{10.00\\\textcolor{gray}{3.90}} & \makecell{15.56\\\textcolor{gray}{11.80}} & \makecell{3.39\\\textcolor{gray}{0.40}} & \makecell{6.64\\\textcolor{gray}{1.10}} & \makecell{2.83\\\textcolor{gray}{0.75}} & 
\makecell{15.69\\\textcolor{gray}{9.92}} & \makecell{6.88\\\textcolor{gray}{3.53}} \\
\midrule  
\includegraphics[width=0.4cm]{internvl.png} InternVL2.5\cite{chen2024expanding} & 8B & \makecell{0.26\\\textcolor{gray}{0.00}} & \makecell{0.38\\\textcolor{gray}{0.00}} & \makecell{13.52\\\textcolor{gray}{3.56}} & \makecell{20.82\\\textcolor{gray}{13.80}} & \makecell{1.96\\\textcolor{gray}{0.00}} & \makecell{5.25\\\textcolor{gray}{0.00}} & \makecell{4.70\\\textcolor{gray}{0.85}} & 
\makecell{9.56\\\textcolor{gray}{3.44}} & \makecell{7.04\\\textcolor{gray}{2.56}} \\
\cmidrule(lr){2-11}
\includegraphics[width=0.4cm]{internvl.png} InternVL2.5\cite{chen2024expanding} & 78B & \makecell{0.24\\\textcolor{gray}{0.10}} & \makecell{0.32\\\textcolor{gray}{0.10}} & \makecell{9.16\\\textcolor{gray}{2.08}} & \makecell{16.18\\\textcolor{gray}{10.00}} & \makecell{4.32\\\textcolor{gray}{0.50}} & \makecell{11.86\\\textcolor{gray}{2.30}} & \makecell{5.48\\\textcolor{gray}{1.25}} & 
\makecell{10.67\\\textcolor{gray}{4.52}} & \makecell{7.29\\\textcolor{gray}{2.55}} \\
\midrule  
\includegraphics[width=0.4cm]{internvl.png} InternVL3\cite{zhu2025internvl3} & 8B & \makecell{1.07\\\textcolor{gray}{0.30}} & \makecell{1.20\\\textcolor{gray}{0.00}} & \makecell{14.36\\\textcolor{gray}{4.42}} & \makecell{13.30\\\textcolor{gray}{6.50}} & \makecell{6.43\\\textcolor{gray}{0.90}} & \makecell{18.73\\\textcolor{gray}{4.60}} & \makecell{4.73\\\textcolor{gray}{1.15}} &
\makecell{15.16\\\textcolor{gray}{7.42}} & 
\makecell{9.26\\\textcolor{gray}{3.19}} \\
\cmidrule(lr){2-11}
\includegraphics[width=0.4cm]{internvl.png} InternVL3\cite{zhu2025internvl3} & 14B & \makecell{0.66\\\textcolor{gray}{0.00}} & \makecell{0.71\\\textcolor{gray}{0.00}} & \makecell{13.24\\\textcolor{gray}{5.31}} & \makecell{19.77\\\textcolor{gray}{13.00}} & \makecell{8.60\\\textcolor{gray}{2.10}} & \makecell{13.17\\\textcolor{gray}{2.40}} & \makecell{10.87\\\textcolor{gray}{3.70}} & 
\makecell{14.57\\\textcolor{gray}{7.76}} & \makecell{10.53\\\textcolor{gray}{4.41}} \\
\cmidrule(lr){2-11}
\includegraphics[width=0.4cm]{internvl.png} InternVL3\cite{zhu2025internvl3} & 38B & \makecell{0.98\\\textcolor{gray}{0.10}} & \makecell{1.76\\\textcolor{gray}{0.20}} & \makecell{12.99\\\textcolor{gray}{4.79}} & \makecell{19.27\\\textcolor{gray}{13.60}} & \makecell{7.63\\\textcolor{gray}{2.10}} & \makecell{17.76\\\textcolor{gray}{2.90}} & 
\makecell{6.47\\\textcolor{gray}{1.75}} & \makecell{16.59\\\textcolor{gray}{10.05}} & \makecell{10.37\\\textcolor{gray}{4.44}} \\
\cmidrule(lr){2-11}
\includegraphics[width=0.4cm]{internvl.png} InternVL3\cite{zhu2025internvl3} & 78B & \makecell{0.20\\\textcolor{gray}{0.00}} & \makecell{0.53\\\textcolor{gray}{0.00}} & \makecell{6.35\\\textcolor{gray}{2.43}} & \makecell{13.03\\\textcolor{gray}{8.00}} & \makecell{3.57\\\textcolor{gray}{0.90}} & \makecell{11.81\\\textcolor{gray}{2.50}} & 
\makecell{3.34\\\textcolor{gray}{0.85}} & \makecell{12.76\\\textcolor{gray}{8.10}} & \makecell{6.44\\\textcolor{gray}{2.90}} \\
\midrule 
\includegraphics[width=0.4cm]{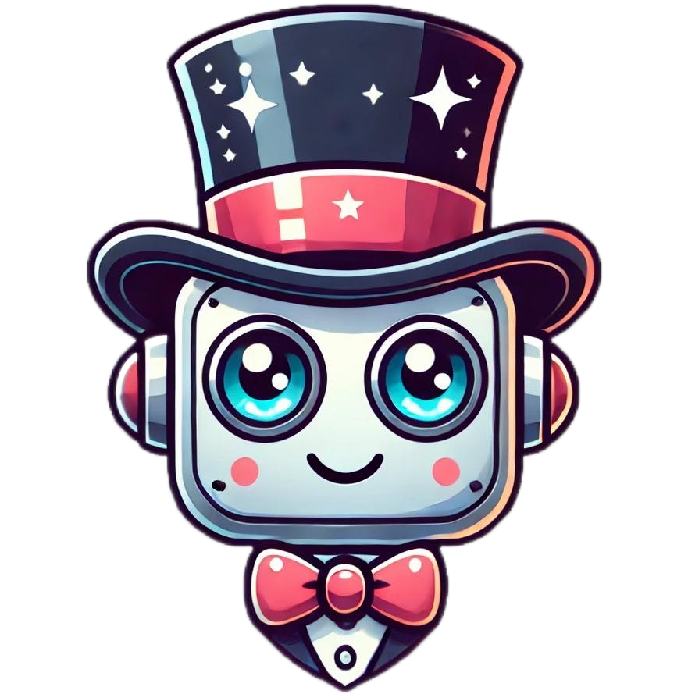} Migician\cite{li2025migician} & 7B & \makecell{\underline{15.26}\\\textcolor{gray}{\underline{7.80}}} & \makecell{\underline{14.49}\\\textcolor{gray}{\underline{6.10}}} & \makecell{18.16\\\textcolor{gray}{7.84}} & \makecell{21.38\\\textcolor{gray}{14.90}} & \makecell{\underline{14.23}\\\textcolor{gray}{\underline{7.20}}} & \makecell{28.87\\\textcolor{gray}{13.70}} & \makecell{\underline{21.41}\\\textcolor{gray}{\underline{12.15}}} &
\makecell{\underline{25.30}\\\textcolor{gray}{\underline{18.02}}} & \makecell{\underline{20.29}\\\textcolor{gray}{\underline{11.39}}} \\    
        \midrule
        \rowcolor[RGB]{234, 238, 234}
        \multicolumn{11}{c}{\bf Medical-domain specialized MLLMs} \\
        \midrule
\includegraphics[width=0.4cm]{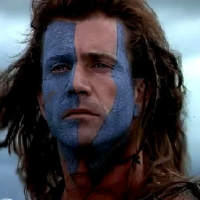} HuatuoGPT-Vision\cite{chen2024huatuogpt} & 7B & \makecell{1.35\\\textcolor{gray}{0.00}} & \makecell{1.84\\\textcolor{gray}{0.20}} & \makecell{10.42\\\textcolor{gray}{2.78}} & \makecell{14.57\\\textcolor{gray}{9.20}} & \makecell{7.99\\\textcolor{gray}{0.80}} & \makecell{15.52\\\textcolor{gray}{2.30}} & \makecell{9.46\\\textcolor{gray}{2.15}} &
\makecell{9.60\\\textcolor{gray}{1.82}} & \makecell{8.97\\\textcolor{gray}{2.36}} \\
\cmidrule(lr){2-11}
\includegraphics[width=0.4cm]{huatuo.png} HuatuoGPT-Vision\cite{chen2024huatuogpt} & 34B & \makecell{1.44\\\textcolor{gray}{0.00}} & \makecell{2.15\\\textcolor{gray}{0.00}} & \makecell{9.41\\\textcolor{gray}{1.65}} & \makecell{13.25\\\textcolor{gray}{8.30}} & \makecell{6.43\\\textcolor{gray}{0.70}} & \makecell{14.53\\\textcolor{gray}{1.40}} & \makecell{10.60\\\textcolor{gray}{2.60}} & 
\makecell{8.60\\\textcolor{gray}{1.75}} & \makecell{8.57\\\textcolor{gray}{2.09}} \\
\midrule
\rowcolor[RGB]{234, 238, 234}
\includegraphics[width=0.4cm]{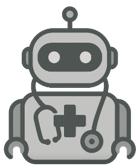} MedSeq-Grounder (Ours) & 7B & \makecell{\textbf{83.29}\\\textcolor{gray}{\textbf{93.20}}} & \makecell{\textbf{83.72}\\\textcolor{gray}{\textbf{94.10}}} & \makecell{\textbf{55.03}\\\textcolor{gray}{\textbf{60.19}}} & \makecell{\textbf{62.10}\\\textcolor{gray}{\textbf{67.20}}} & \makecell{\textbf{74.11}\\\textcolor{gray}{\textbf{82.60}}} & \makecell{\textbf{85.25}\\\textcolor{gray}{\textbf{98.80}}} & \makecell{\textbf{78.77}\\\textcolor{gray}{\textbf{82.75}}} & 
 \makecell{\textbf{60.43}\\\textcolor{gray}{\textbf{65.59}}} & \makecell{\textbf{72.55}\\\textcolor{gray}{\textbf{79.71}}} \\
        \bottomrule
    \end{tabular}
    \label{tab:main_results}
\end{table}

\section{Experiments}
\label{others}

\subsection{Experiment setup}

In this study, we evaluate model performance under a zero-shot setting, where the models were prompted to perform inference without access to in-context examples. We use average Intersection over Union (IoU) and ACC@0.5 as the evaluation metric.

\subsection{Models}
We benchmark a diverse collection of state-of-the-art MLLMs on MedSG-Bench, including 1) general-purpose models that have extended capabilities in the medical domain, and 2) medical-domain specialized models that are meticulously trained for clinical medicine. All models support image sequence input and span parameter scales from approximately 3 billion to 70 billion. We use publicly released checkpoints from their official Hugging Face repositories\cite{jain2022hugging} and, by default, select the latest or best-performing version within each model family.

\paragraph{General-Purpose MLLMs} We evaluate Qwen2.5-VL (3B, 7B, 32B, 72B)\cite{bai2025qwen2}, MiniCPM-V-2\_6\cite{yao2024minicpm}, MiniCPM-O-2\_6\cite{minicpm-o}, mPlug-owl3\cite{ye2024mplug}, Mantis-Idefics2\cite{jiang2024mantis}, llava\_onevision (7B, 72B)\cite{li2024llava}, internvl2 (8B, 78B)\cite{chen2024far,chen2024internvl}, internvl2\_5 (8B, 78B)\cite{chen2024expanding}, internvl3 (8B, 14B, 38B, 78B)\cite{zhu2025internvl3}. For grounding-oriented MLLMs, we evaluate Migician\cite{li2025migician}, which supports free-form multi-image grounding and has strong instruction-following capability.
\paragraph{Medical-domain specialized MLLMs} we evaluate HuatuoGPT-Vision (7B, 34B)\cite{chen2024huatuogpt}, which is built on a large-scale and high-quality medical VQA dataset, PubMedVision.

\subsection{Main Results}

Based on the evaluation results presented in Table~\ref{tab:main_results}, we have some findings as follows:
\paragraph{Grounding in medical image sequences is still challenging for all MLLMs} Our MedSG-Bench provides a comprehensive multitask challenge, revealing that even the top-performing model Migician is limited to the average IoU of 20.29\% and Acc@0.5 of 11.39\% in zero-shot setting. In particular, most MLLMs struggle with the Image Difference Grounding task. Moreover, the most advanced models do not consistently excel across all tasks, for example, while migician achieves relatively high accuracy on the cross-modal grounding task, its performance on multi-view grounding or object tracking remains notably lower than Mantis, highlighting the challenge of generalization across diverse grounding scenarios. With instruction tuning on our MedSeqVG-188K dataset, the proposed MedSeq-Grounder achieves state-of-the-art performance across all tasks, demonstrating its effectiveness and robustness in sequential medical visual grounding.

\paragraph{Medical-domain specialized models are often worse than general-purpose models} While specialist models are explicitly developed for the medical domain, they often underperform non-specialist open-source models. For example, HuatuoGPT-Vision-7B, lags behind Qwen2.5-VL-7B by 3.34\% in average IoU and 2.54\% in Acc@0.5 on MedSG-Bench. Notably, it even performs worse than the smaller-sized Qwen2.5-VL-3B model. This performance gap may be attributed to the nature of training data used for domain adaptation. Most existing medical instruction-tuning datasets focus predominantly on image-level understanding tasks, such as classification or report summarization. While HuatuoGPT-Vision is built upon Qwen-VL, its further tuning on understanding-centric medical data appears to have degraded its grounding capability. This reflects a case of catastrophic forgetting, where the model’s original ability for spatial alignment is compromised due to continued learning on tasks that lack grounding supervision.

\paragraph{Larger or newer models do not guarantee improved grounding performance} Although model scale and recency are commonly associated with improved performance, we find that larger or more recently released models do not necessarily exhibit stronger grounding capabilities in medical image sequences. For instance, InternVL2.5-8B and InternVL3-8B both underperform compared to the earlier InternVL2-8B model, despite architectural updates and increased pretraining. Similarly, MiniCPM-O-2\_6 lags behind MiniCPM-V-2\_6, highlighting that newer instruction-tuned variants may sacrifice grounding performance in favor of improvements on general-purpose understanding tasks. In some cases, such as with the InternVL family, even the 70B-scale model yields worse results on MedSG-Bench compared to its 8B counterpart, indicating that grounding ability may not scale proportionally with model size. These results suggest that many recent models are primarily optimized for high-level semantic tasks, such as open-ended QA or captioning, and are trained on instruction-tuning datasets that provide little to no supervision for spatial localization or visual grounding. This observation further underscores the importance of dedicated benchmarks like MedSG-Bench, which are specifically designed to evaluate fine-grained grounding and spatial alignment across sequential medical images.

\section{Conclusion}
This work introduces MedSG-Bench, the first benchmark specifically designed to evaluate the fine-grained visual grounding capabilities of MLLMs in sequential medical images. Through systematic evaluations on eight clinically inspired grounding tasks, we find that all current MLLMs exhibit substantial limitations in medical image sequences grounding. To address these challenges, we construct a grounding instruction-tuning dataset, MedSG-188K, and develop MedSeq-Grounder. We hope our benchmark, dataset, and model will together advance the development of visual grounding in medical image sequences.

\bibliographystyle{plain}
\bibliography{neurips_2025}

\newpage
\appendix

\section{Dataset Details}

In this section, we provide the detailed datasets used in MedSG-Bench, including the name of the dataset, the modality, the dimension of data, and the accessible links. As shown in Table~\ref{tab:dataset_statistics}, MedSG-Bench is constructed from 76 datasets across 10 medical image modalities.

\small
\begin{longtable}{p{0.18\textwidth}|p{0.11\textwidth}|p{0.04\textwidth}|p{0.6\textwidth}}

\caption{Detailed datasets information in MedSG-Bench. }
\label{tab:dataset_statistics} \\
    \hline
    Dataset & Modality & Dim & Accessible links \\
    \hline
    \endfirsthead

    \endhead
    \hline
    \endfoot
    \hline
    \endlastfoot


 4C2021\cite{4c2021} & CT & 3D & \url{https://aistudio.baidu.com/datasetdetail/89548} \\
 \hline
 AbdomenCT1K\cite{ma2021abdomenct} & CT & 3D & \url{https://github.com/JunMa11/AbdomenCT-1K } \\
 \hline
 ACDC\cite{bernard2018deep} & MRI & 3D & \url{https://humanheart-project.creatis.insa-lyon.fr/database/} \\
 \hline
 AMOS22\cite{ji2022amos} & CT, MRI & 3D & \url{https://amos22.grand-challenge.org/} \\
 \hline
 ATM22\cite{zhang2023multi} & CT & 3D & \url{https://atm22.grand-challenge.org/} \\
 \hline
 Atria Segmentation\cite{xiong2021global} & MRI & 3D & \url{https://www.cardiacatlas.org/atriaseg2018-challenge/atria-seg-data/} \\
 \hline
 AutoLaparo\cite{wang2022autolaparo} & Colonoscopy & 2D & \url{https://autolaparo.github.io/ } \\
 \hline
 BAGLS\cite{gomez2020bagls} & Endoscopy & 2D & \url{https://www.kaggle.com/datasets/gomezp/benchmark-for-automatic-glottis-segmentation} \\
 \hline
 BraimMRI\cite{nickparvar2021brain} & MRI & 3D & \url{https://www.kaggle.com/datasets/masoudnickparvar/brain-tumor-mri-dataset} \\
 \hline
 BrainPTM\cite{avital2019neural}\cite{nelkenbaum2020automatic} & MRI & 3D & \url{https://brainptm-2021.grand-challenge.org/} \\
 \hline
 \makecell{BraTS2020\cite{menze2014multimodal}\cite{bakas2017advancing}\\\cite{bakas2018identifying}} & MRI & 3D & \url{https://service.tib.eu/ldmservice/dataset/brats2020} \\
 \hline
 BUSI\cite{al2020dataset} & US & 2D & \url{https://scholar.cu.edu.eg/?q=afahmy/pages/dataset} \\
 \hline
 CAD-PE\cite{gonzalez2020computer} & CT & 3D & \url{https://ieee-dataport.org/open-access/cad-pe} \\
 \hline
 CAMUS\cite{leclerc2019deep} & US & 2D & \url{https://www.creatis.insa-lyon.fr/Challenge/camus/} \\
 \hline
 Cause07\cite{van20073d} & MRI & 3D & \url{https://cause07.grand-challenge.org/} \\
 \hline
 CBCT3D\cite{cui2022ctooth+}\cite{cui2022ctooth} & CBCT & 3D & \url{https://toothfairy.grand-challenge.org/} \\
 \hline
 Chestimage\cite{Chestimage} & X-Ray & 2D & \url{https://tianchi.aliyun.com/dataset/83075} \\
 \hline
 CMRxMotions\cite{wang2022extreme} & MRI & 3D & \url{https://www.synapse.org/Synapse:syn28503327/} \\
 \hline
 COVID-19\cite{juanying2022xr} & CT & 3D & \url{https://medicalsegmentation.com/covid19/} \\
 \hline
 COVID19CTscans\cite{jun2020covid} & CT & 3D & \url{https://zenodo.org/records/3757476} \\
 \hline
 COVID-19-20\cite{roth2022rapid} & CT & 3D & \url{https://covid-segmentation.grand-challenge.org/} \\
 \hline
 Covid19cxr\cite{cohen2020covid} & X-ray & 2D & \url{https://github.com/ieee8023/covid-chestxray-dataset} \\
 \hline
 Cranium\cite{hssayeni2020computed} & CT & 3D & \url{https://tianchi.aliyun.com/dataset/82967} \\
 \hline
 CT-ORG\cite{rister2020ct} & CT & 3D & \url{https://www.cancerimagingarchive.net/collection/ct-org/} \\
 \hline
 CTSpine1K\cite{deng2021ctspine1k} & CT & 3D & \url{https://github.com/MIRACLE-Center/CTSpine1K} \\
 \hline
 CVC-ClinicDB\cite{bernal2015wm} & Colonoscopy & 2D & \url{https://polyp.grand-challenge.org/CVCClinicDB/} \\
 \hline
 DRISHTI-GS\cite{sivaswamy2014drishti} & Fundus & 2D & \url{https://www.kaggle.com/datasets/lokeshsaipureddi/drishtigs-retina-dataset-for-onh-segmentation} \\
 \hline
 EMIDEC\cite{lalande2022deep} & MRI & 3D & \url{https://emidec.com/dataset} \\
 \hline
 EndoTect2020\cite{hicks2021endotect} & Colonoscopy & 2D & \url{https://osf.io/mh9sj/} \\
 \hline
 EndoVis15\cite{bernal2017comparative} & Colonoscopy & 2D & \url{https://endovis.grand-challenge.org/} \\
 \hline
 EndoVis2017\cite{allan20192017} & Colonoscopy & 2D & \url{https://endovissub2017-roboticinstrumentsegmentation.grand-challenge.org/} \\
 \hline
 GAMMA\cite{fu2018joint}\cite{fu2020age}\cite{orlando2020refuge} & Fundus & 2D & \url{https://gamma.grand-challenge.org/Home/} \\
 \hline
 HaN-Seg\cite{podobnik2023han} & CT, MRI & 3D & \url{https://zenodo.org/records/7442914} \\
 \hline
 Hvsmr2016\cite{pace2015interactive} & MRI & 3D & \url{http://segchd.csail.mit.edu/data.html} \\
 \hline
 I2CVB\cite{lemaitre2015computer} & MRI & 3D & \url{https://i2cvb.github.io/} \\
 \hline
 InSTANCE2022\cite{li2023state}\cite{li2021hematoma} & CT & 3D & \url{https://instance.grand-challenge.org/} \\
 \hline
 iseg2017\cite{wang2019benchmark} & MRI & 3D & \url{https://iseg2017.web.unc.edu/download/} \\
 \hline
 ISIC2018\cite{codella2019skin}\cite{tschandl2018ham10000} & Dermoscopy & 2D & \url{https://challenge.isic-archive.com/data/#2018} \\
 \hline
 ISLES-ATLAS\cite{hernandez2022isles} & MRI & 3D & \url{https://atlas.grand-challenge.org/} \\
 \hline
 ISLES-MM\cite{hernandez2022isles} & MRI & 3D & \url{https://isles22.grand-challenge.org/} \\
 \hline
 JSRT\cite{shiraishi2000development} & X-ray & 2D & \url{http://imgcom.jsrt.or.jp/minijsrtdb/} \\
 \hline
 KvasirInstrument\cite{jha2021kvasir} & Colonoscopy & 2D & \url{https://datasets.simula.no/kvasir-instrument/} \\
 \hline
 LMSLS\cite{carass2017longitudinal} & MRI & 3D & \url{https://smart-stats-tools.org/lesion-challenge-2015} \\
 \hline
 LUNA16\cite{setio2017validation} & CT & 3D & \url{https://luna16.grand-challenge.org/Download/} \\
 \hline
\makecell{MMWHS\cite{gao2023bayeseg}\cite{luo2022mathcal}\\\cite{wu2022minimizing}\cite{zhuang2018multivariate}} & CT, MRI & 3D & \url{https://www.ub.edu/mnms/} \\
 \hline
 MRSpineSeg\cite{pang2022dgmsnet}\cite{pang2020spineparsenet} & MRI & 3D & \url{https://mosmed.ai/datasets/covid19_1110} \\
 \hline
 MSD02\cite{simpson2019large} & MRI & 3D & \url{http://medicaldecathlon.com/} \\
 \hline
 MSD04\cite{antonelli2022medical} & MRI & 3D & \url{http://medicaldecathlon.com/} \\
 \hline
 MSD05\cite{antonelli2022medical} & MRI & 3D & \url{http://medicaldecathlon.com/} \\
 \hline
 \makecell{MyoPS2020\cite{gao2023bayeseg}\cite{luo2022mathcal}\\\cite{zhuang2018multivariate}} & MRI & 3D & \url{https://zmiclab.github.io/zxh/0/myops20/} \\
 \hline
 NCI-ISBI2013\cite{NCI-ISBI} & MRI & 3D & \url{https://www.cancerimagingarchive.net/analysis-result/isbi-mr-prostate-2013/} \\
 \hline
PadChest\cite{bustos2020padchest} & X-ray & 2D & \url{https://bimcv.cipf.es/bimcv-projects/padchest/} \\
 \hline
 PALM\cite{fu2019palm} & Fundus & 2D & \url{https://ieee-dataport.org/documents/palm-pathologic-myopia-challenge} \\
 \hline
 Parse2022\cite{luo2023efficient} & CT & 3D & \url{https://parse2022.grand-challenge.org/Dataset/} \\
 \hline
 PCXA\cite{candemir2013lung}\cite{jaeger2013automatic} & X-ray & 2D & \url{https://lhncbc.nlm.nih.gov/LHC-downloads/downloads.html} \\
 \hline
 PDDCA\cite{raudaschl2017evaluation} & CT & 3D & \url{https://www.imagenglab.com/newsite/pddca/} \\
 \hline
 Pelvic1K\cite{liu2021deep} & CT & 3D & \url{https://zenodo.org/record/4588403} \\
 \hline
 Promise09\cite{dowling2009automatic} & MRI & 3D & \url{https://www.na-mic.org/wiki/Training_Data_Prostate_Segmentation_Challenge_MICCAI09} \\
 \hline
 PROMISE12\cite{litjens2014evaluation} & MRI & 3D & \url{https://zenodo.org/records/8026660} \\
 \hline
 \makecell[l]{QaTa-COV19\cite{degerli2022osegnet}\\
\cite{ahishali2021advance}\cite{degerli2021covid}\cite{degerli2021reliable}\cite{yamac2021convolutional}} & X-ray & 2D & \url{https://www.kaggle.com/datasets/aysendegerli/qatacov19-dataset} \\
 \hline
 QUBIQ2020\cite{li2024qubiq} & CT & 2D & \url{https://qubiq.grand-challenge.org/} \\
 \hline
 REFUGE\cite{orlando2020refuge}\cite{li2020development} & Fundus & 2D & \url{https://refuge.grand-challenge.org/} \\
 \hline
 RIGA+\cite{hu2022domain} & Fundus & 2D & \url{https://zenodo.org/records/6325549} \\
 \hline
 RIM\_ONE\cite{fumero2011rim} & Fundus & 2D & \url{https://github.com/miag-ull/rim-one-dl} \\
 \hline
 SegRap2023\cite{luo2025segrap2023} & CT & 2D & \url{https://segrap2023.grand-challenge.org/dataset/} \\
 \hline
 SegTHOR\cite{lambert2020segthor} & CT & 3D & \url{https://competitions.codalab.org/competitions/21145} \\
 \hline
 SIIM-ACR\cite{zawacki2019siim} & X-ray & 2D & \url{https://www.kaggle.com/c/siim-acr-pneumothorax-segmentation} \\
 \hline
 SKI10\cite{heimann2010segmentation} & MRI & 3D & \url{https://ski10.grand-challenge.org/} \\
 \hline
 SLAWT\cite{karim2018algorithms} & MRI & 3D & \url{http://stacom.cardiacatlas.org/} \\
 \hline
 TBAD\cite{yao2021imagetbad} & CTA & 3D & \url{https://www.kaggle.com/datasets/xiaoweixumedicalai/imagetbad} \\
 \hline
 TN-SCUI\cite{gireesha2014thyroid} & US & 2D & \url{https://tn-scui2020.grand-challenge.org/} \\
 \hline
 VESSEL12\cite{rudyanto2014comparing} & CT & 3D & \url{https://vessel12.grand-challenge.org/} \\
 \hline
 VINDR-Mammo\cite{nguyen2023vindr} & X-ray & 2D & \url{https://www.physionet.org/content/vindr-mammo/1.0.0/} \\
 \hline
 Verse19\cite{loffler2020vertebral}\cite{sekuboyina2021verse} & CT & 3D & \url{https://github.com/anjany/verse} \\
 \hline
 WMH\cite{kuijf2019standardized} & MRI & 3D & \url{https://dataverse.nl/dataset.xhtml?persistentId=doi:10.34894/AECRSD} \\
 \hline
 WORD\cite{luo2022word} & CT & 3D & \url{https://github.com/HiLab-git/WORD} \\

\end{longtable}

\newpage

\clearpage
\newpage
\section{Data statistics of MedSG-188K}
\label{datades2}

\begin{figure}[h]
  \centering
  \includegraphics[width=\textwidth]{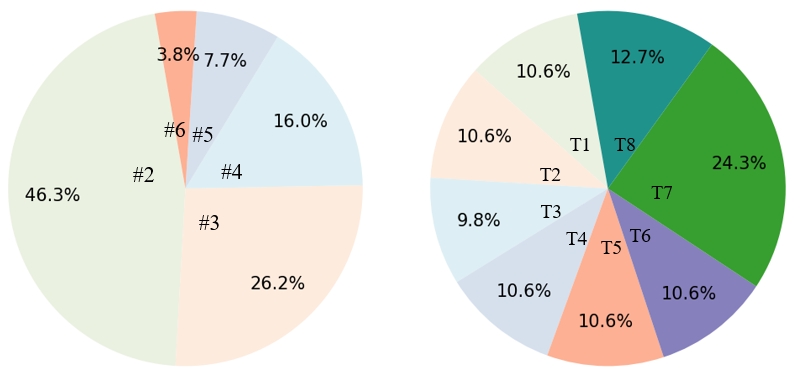}
  \caption{Proportions of image sequence length (\textbf{left}), data distribution across tasks (\textbf{right}) in MedSG-188K.}
  \label{fig:datades}
\end{figure}

\section{Evaluation Metric}
We evaluate model performance using two standard metrics: Intersection over Union (IoU) and Accuracy at IoU threshold 0.5 (Acc@0.5). These metrics are widely adopted in visual grounding to measure localization quality.

IoU quantifies the overlap between the predicted bounding box $B_{\text{pred}}$ and the ground-truth bounding box $B_{\text{gt}}$, and is defined as:

\begin{equation}
\mathrm{IoU} = \frac{\mathrm{Area}(B_{\text{pred}} \cap B_{\text{gt}})}{\mathrm{Area}(B_{\text{pred}} \cup B_{\text{gt}})}
\end{equation}
Acc@0.5 measures the proportion of predictions whose IoU with the ground truth exceeds 0.5. It is defined as:
\begin{equation}
\mathrm{Acc@0.5} = \frac{1}{N} \sum_{i=1}^{N} \mathbb{I}(\mathrm{IoU}_i \geq 0.5)
\end{equation}
Here, \( N \) is the total number of samples, and \( \mathbb{I}(\cdot) \) is the indicator function that returns 1 if the condition is true, and 0 otherwise.

\end{document}